\documentclass[10pt,twocolumn,letterpaper]{article}
\usepackage{xcolor}
\definecolor{hierar}{RGB}{178, 9, 78}
\definecolor{m2t}{RGB}{153, 79, 13}
\definecolor{m+t}{RGB}{183, 132, 6}
\definecolor{editing}{RGB}{2, 96, 11}
\definecolor{input}{RGB}{143, 105, 0}
\definecolor{model}{RGB}{181, 26, 0}
\definecolor{output}{RGB}{156, 38, 78}
\definecolor{loss}{RGB}{0, 76, 153}

\usepackage[pagenumbers]{cvpr}              %

\newcommand{\methodname}[0]{UniMotion}

\newcommand{\x}{\mathbf{x}}
\newcommand{\y}{\mathbf{y}}

\renewcommand{\paragraph}[1]{{\vspace{0.6mm}\noindent \bf #1}.}

\newcommand{\PAR}[1]{\vskip4pt \noindent{\bf #1~}}

\usepackage{multirow}
\usepackage{booktabs} %
\usepackage{algorithm}
\usepackage{algpseudocode}
\usepackage{amsmath}

\definecolor{cvprblue}{rgb}{0.21,0.49,0.74}
\usepackage[pagebackref,breaklinks,colorlinks,citecolor=cvprblue]{hyperref}

\def\paperID{262} %
\def\confName{3DV\xspace}
\def\confYear{2025\xspace}

\title{\methodname{}: Unifying 3D Human Motion Synthesis and Understanding}

\author{
\begin{tabular}{cccccccccccc}\multicolumn{3}{c}{Chuqiao Li \textsuperscript{1}} & \multicolumn{3}{c}{Julian Chibane \textsuperscript{1,2}} & \multicolumn{3}{c}{Yannan He \textsuperscript{1}}& \multicolumn{3}{c}{Naama Pearl \textsuperscript{1}} \\  \multicolumn{6}{c}{Andreas Geiger \textsuperscript{1}} & \multicolumn{6}{c}{Gerard Pons-Moll\textsuperscript{1,2}} \end{tabular}\\\\
{\small \textsuperscript{1}Tübingen AI Center, University of Tübingen, Germany, \qquad \textsuperscript{2}Max Planck Institute for Informatics, Saarland Informatics Campus, Germany}  \\
{\tt\scriptsize \{chuqiao.li, yannan.he, naama.pearl, a.geiger, gerard.pons-moll\}@uni-tuebingen.de,} \vspace{-2mm}\\ {\tt\scriptsize jchibane@mpi-inf.mpg.de}
}

\begin{document}

\twocolumn[{%
\vspace{-12mm}
\renewcommand\twocolumn[1][]{#1}%
\maketitle
\vspace{-3mm}
\begin{center}
    \centering
    \captionsetup{type=figure}
    \vspace{-18pt}
    \includegraphics[width=\textwidth]{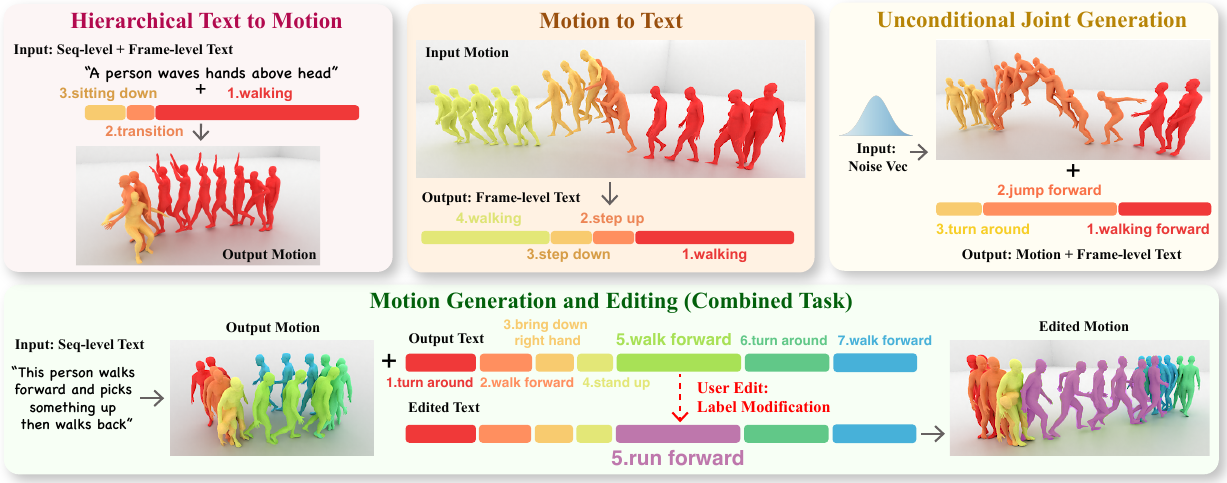}
    \vspace{-4mm}
    \caption{\textbf{Universality of \methodname{}.}
    Our model can generate motion from compositional sequence- and frame-level text (\textbf{\textcolor{hierar}{Hierachical Text to Motion}}), generate detailed per frame motion descriptions (\textbf{\textcolor{m2t}{Motion to Text}}), generate motion and accurate frame-level text descriptions from noise (\textbf{\textcolor{m+t}{Unconditional Joint Generation}}), amongst other use cases outlined in our experiments section.
    Tasks can be combined for a controllable generation: users can generate motion from a coarse sentence, our model additionally generates detailed text descriptions, which can be edited and used for regeneration, generating the desired edited motion (\textbf{\textcolor{editing}{Motion Generation and Editing}}).
    }
    \label{fig:teaser}
\end{center}
}]

\begin{abstract}
We introduce \methodname{}, the first unified multi-task human motion model capable of both flexible motion control and frame-level motion understanding. 
While existing works control avatar motion with global text conditioning, or with fine-grained per frame scripts, none can do both at once.
In addition, none of the existing works can output frame-level text paired with the generated poses.
In contrast, \methodname{} allows to control motion with global text, or local frame-level text, or both at once, providing more flexible control for users.
Importantly, \methodname{} is the first model which by design outputs local text paired with the generated poses, allowing users to know what motion happens and when, which is necessary for a wide range of applications.
We show \methodname{} opens up new applications: 1.) hierarchical control, allowing users to specify motion at different levels of detail,
2.) obtaining motion text descriptions for existing MoCap data or youtube videos
3.) allowing for editability, generating motion from text and editing the motion via text edits. 
Moreover, \methodname{} attains state-of-the-art results for the frame-level text-to-motion task on the established HumanML3D dataset.
The pre-trained model and code are available available on our project page at \url{https://coral79.github.io/uni-motion/}.
    \vspace{-0.3cm}

\end{abstract}

\section{Introduction}
\label{sec:intro}

Human motion synthesis is important for gaming, robotics and AR/VR applications. In real-world scenarios, avatars need to be controlled at multiple levels of abstraction.
Effective controllability requires that an avatar be capable of executing detailed local sub-tasks according to a timeline while simultaneously understanding the overall global objective.
In addition, the synthesis model should be aware of what action happens and when -- an essential feature of biological intelligence to react to the external world.

However, current motion synthesis methods focus on either global per sequence-level control, or local per frame-level control, but don't allow for both. This results in single-level conditioning, thereby \textbf{lacking hierarchical control.} 
Importantly, these models also lack fine-grained motion awareness, specifically, the ability to output motion descriptions for each pose in the generated output motion sequence.
The frame-level text-to-motion methods ~\cite{TEACH:3DV:2022,barquero2024seamless,shafir2023human} provide detailed manipulation of individual frames. However, it can be impractical to specify the exact duration of each action in some situations, and ensuring overall semantic plausibility throughout the entire sequence remains challenging for these models.
Conversely, the sequence-level text-to-motion methods~\cite{tevet2023human,zhou2024avatargpt,jiang2024motiongpt} focus on achieving natural overall motion but struggles with fine-grained control. %
Furthermore, current models lack semantic awareness of the synthesized motion -- there is no understanding of what action occurs when. 
Thus, they are \textbf{lacking motion understanding}, which is crucial for reacting to the external world and allows for action-specific editing in animation applications.
While some works have made progress in this direction \cite{zhou2024avatargpt,jiang2024motiongpt} by predicting sequence-level text descriptions from motion, they fail to provide fine-grained frame-level text. Overall, despite their potential synergies, motion understanding and synthesis have been treated in isolation in the literature.

In this paper, we introduce \methodname{}, the first unified multi-task model capable of both flexible motion control and frame-level motion understanding.
\methodname{} takes as input, global sequence level or local frame-level text inputs or human motion sequences, or any subsets thereof, or no input in case of unconditional generation.
The output of our model is either fine grained, per pose text descriptions, or human motion sequences.
This flexibility, allows us to train our model from different data sources.
Moreover, by design, we unify tasks that are usually treated in separation by prior works, such as \textit{Frame-Level Text-to-Motion},\textit{ Sequence-Level Text-to-Motion} and \textit{Motion-to-Text}, into a single simple unified model, trained a single time. 
Importantly, \methodname{}'s flexibility also allows for novel tasks not previously considered by prior work like 1.) unconditional generation of human motion with corresponding frame-level text descriptions and 2.) generation of frame-level text from motion, providing granular, time-aware annotations (see Fig.~\ref{fig:teaser} for an illustration our diverse tasks). 

To accomplish this, our model utilizes a transformer architecture with temporal alignment between the motion and frame-level text.
We further enhance this by diffusing the local text together with the poses, using different diffusion time variables for each, inspired by the approach in Unidiffuser~\cite{bao2023one}. Specifically, the local text is tokenized and frame-wise aligned with the 3D poses, while the global text is injected as a global token.
This design allows \methodname{} to dynamically switch between global, local, or combined conditioning signals at test time, providing flexibility in motion generation and understanding.
During training, we sample from all possible distributions (global and/or local conditioning, or unconditional), alternating between providing noise and signal to the model for each modality.
This method effectively teaches the model both unconditional and conditional distributions, equipping it with the ability to handle various inputs.

\textbf{Real-world applicability.} We demonstrate practical utility across various real-world scenarios:

\textit{2D Video Annotation:}
We annotate human motion extracted from YouTube videos with frame-level text, by feeding \methodname{} with human pose estimation (HPE) results. This annotation can serve as close captions for the visually impaired.
\textit{4D Mocap Annotation:}
We annotate human motion captures, e.g. obtained from IMUs, with frame-level text. This provides automated insights and descriptions into the captured motions, e.g. allowing for text search retrieval of motion sequences.
\textit{Hierachical Control:}
We provide examples of generating motion sequences with two levels of abstraction, specifying a general motion for arms via global text, and a fine-grained motion sequence for the rest of the body via local-level text.
\textit{Motion editing:}
We show that \methodname{} can be used for content creation, where controllability of the motion is important. Given a global text description, a user can generate an initial motion including a local-level text description. The user can then edit the motion as desired, editing the text segments and regenerating the motion. In summary, our \textbf{key contributions} are:

\begin{itemize}
    \item \textbf{Unified Synthesis and Understanding}: We introduce \methodname{}, the first unified probabilistic motion model allowing for sampling from the joint and all possible conditionals. It unifies tasks that are usually treated in separation by prior works,
    while also allowing for novel tasks not previously considered.

    \item \textbf{Results and Applications}: 
    We show applicability to 2D Video Annotation, 4D Mocap Annotation, Hierachical Control and Motion editing. Moreover \methodname{} attains state-of-the-art results for the frame-level text-to-motion task on the established HumanML3D dataset. Code and models will be released upon acceptance.
\end{itemize}

\section{Related Work}
\label{sec:related}

\paragraph{Conditional human motion synthesis.} 
Synthesizing human motion has been a long-standing challenge. Recent studies in motion generation have shown notable progress in synthesizing movements conditioned on diverse modalities such as text~\cite{shafir2023human,DBLP:journals/corr/abs-2306-00416,tevet2022motionclip,tevet2023human,petrovich22temos, petrovich2024multi}, music~\cite{aist, li2021learn}, scenes~\cite{wang2022sceneaware,mir23origin}, and interactive objects~\cite{samp,nsm, zhang2022couch,zhang2024force, xiao2023unified, li2023OMOMO, yi2024tesmo, xu2023interdiff}.
Recent years have witnessed substantial advancements in text-driven motion generation~\cite{petrovich22temos, Guo2022CVPR, chuan2022tm2t, zhang2023generating, tevet2022motionclip, motionlcm}. Notably, diffusion-based generative models have emerged as potent tools, exhibiting impressive performance on leading benchmarks for text-to-motion tasks. Pioneering efforts such as MotionDiffuse~\cite{zhang2022motiondiffuse}, MDM~\cite{tevet2023human}, and FLAME~\cite{kim2022flame} represent early applications of diffusion models to text-driven motion generation. Building upon this foundation, MLD~\cite{chen2023executing} further harnesses latent diffusion models, while ReMoDiffuse~\cite{zhang2023remodiffuse} integrates retrieval techniques into the motion generation pipeline. Recent MotionLCM~\cite{motionlcm} accelerates the sampling speed by adopting consistency model in motion latent space. Noteworthy, OmniControl~\cite{xie2024omnicontrol} specializes in fine-grained spatial control of body joints.

\paragraph{Text-to-motion generation models.}
The current landscape of text-to-motion generation models can be categorized into two main streams of controllability: (a) global text-based control and (b) Fine-grained local text-based control. Among the former, MotionGPT~\cite{jiang2024motiongpt} utilizes pre-trained language models and motion-specific vector quantized models to conceptualize human motion as a language. Similarly, AvatarGPT~\cite{zhou2024avatargpt} proposes a top-down approach to address end-to-end motion planning and synthesis.

Conversely, research focusing on short, specific instructions presents another avenue. PriorMDM~\cite{shafir2023human} introduces a two-stage method that synthesizes short motion sequences and their padded transitions. However, due to the lack of effective supervised learning, motions generated by such methods often exhibit artifacts, such as abrupt speed changes. FineMoGen~\cite{zhang2024finemogen} proposes diffusion-based motion generation and editing for fine-grained per-body part motion control, albeit requiring detailed per-body part instructions as input. Closely aligned with our work are methods enabling temporal control of motion, where the length of each motion segment can be controlled at the frame level. FlowMDM~\cite{barquero2024seamless} demonstrates impressive results in seamless transitions between local motion segments, while STMC~\cite{petrovich2024multi} proposes a hybrid method for spatial and temporal motion composition of multi-stream motion using off-the-shelf pre-trained motion models~\cite{tevet2023human}. Notably, these methods do not condition on global text, resulting in a lack of awareness of the global motion context and less natural motion transitions.

Our method combines the advantages of both categories. It is the first method enabling the generation of human motion conditioned both at the abstract level with global text and at the detailed level with local texts.

\paragraph{Human motion understanding.}
Understanding the meaning of human motion has been a long-standing research topic, this has been approached by describing human motion with predefined action labels~\cite{motionbert2022,Chi_2022_CVPR}, which have dominated this field for some time. However, these methods have obvious limitations, they are not appropriate to describe complex motion sequences. Recently, the text annotated motion datasets~\cite{BABEL:CVPR:2021,Guo2022CVPR,cai2022humman} have enabled the methods ~\cite{jiang2024motiongpt,chuan2022tm2t,zhang2023generating} that learn the mutual mapping between human motion sequences and natural language descriptions. While these works produce impressive, they fall short in generating accurate per-frame language descriptions. More recently, methods such as~\cite{huang2024como, goel2024iterative} have achieve motion editing based on more fine-grained conditions, such as per body part condition. However, they still lack the capability for temporal editing. UniMotion is the first approach that not only generates per-frame language descriptions but also allows for motion generation over specified time spans, thus advancing the understanding of human motion.

\begin{figure*}[!ht]
    \centering
    \includegraphics[width=0.85\linewidth]{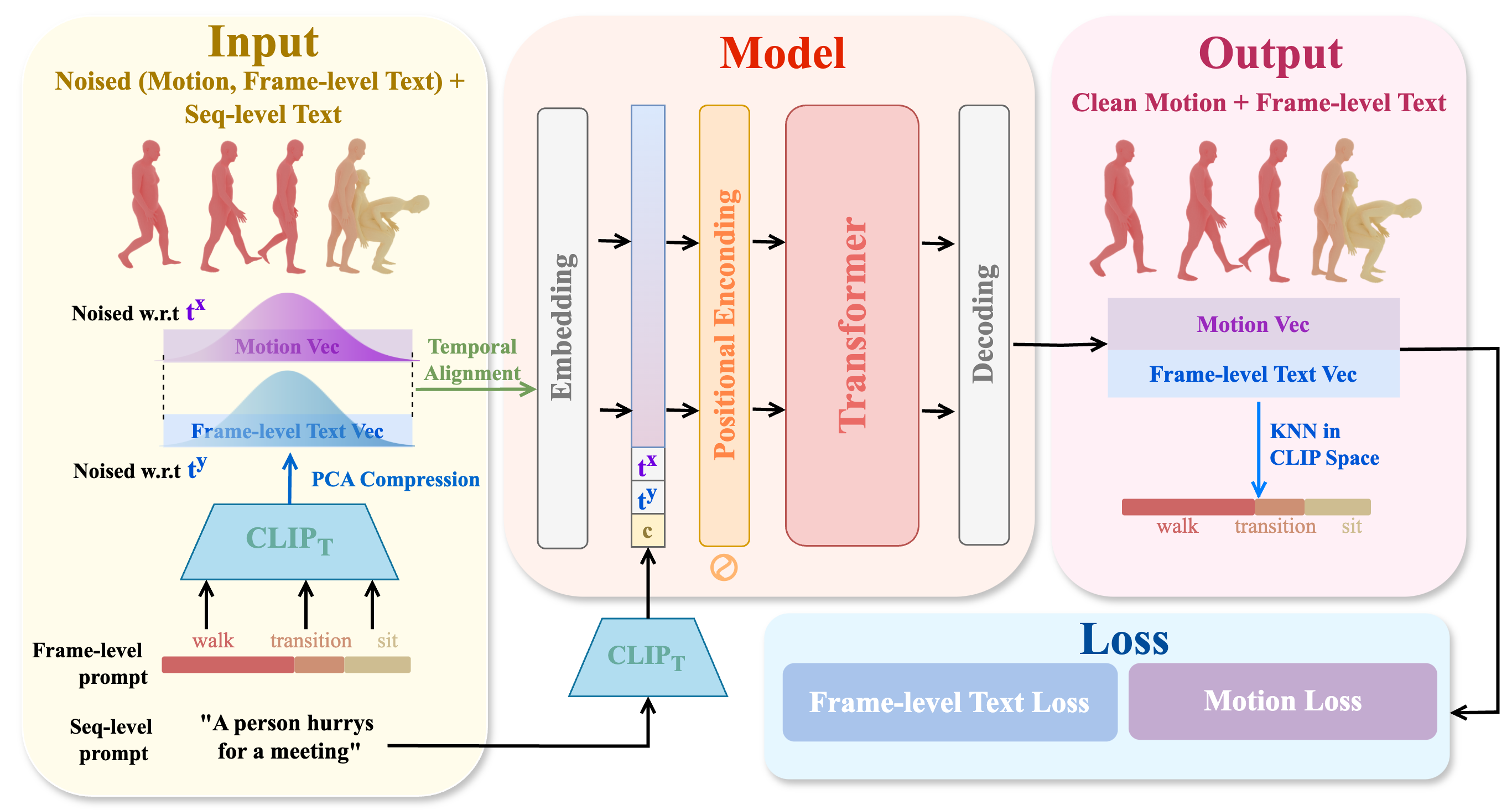}
    \vspace{-5mm}
    \caption{
    \textbf{Overview of \methodname{}}. 
    \methodname{} is a transformer-based diffusion model (\textbf{\textcolor{model}{Model}}) that can be input conditioned on a) human motion, b) clip embedded frame-level text, or c) sequence-level text (\textbf{\textcolor{input}{Input}}) or any subsets thereof or none, and instead supplied with noise.
    At it's core it allows to diffuse motion and text individually, implemented via separate denoising timesteps $t^x$ and $t^y$.
    After training with Frame-level text Losses and Motion losses (\textbf{\textcolor{loss}{Loss}}), see Sec. \ref{subsec:method_mmd}. \methodname{} can output clean, noise-free motion, and frame-level text descriptions explaining the generated motions. (\textbf{\textcolor{output}{Output}})
    }

    \label{fig:overview}
\end{figure*}

\section{Preliminary: Motion Diffusion Model}
\label{sec:background}

We provide a brief overview of the Human Motion Diffusion Model (MDM) ~\cite{tevet2023human}, which is designed for sequence-level text-to-motion synthesis. This model serves as a building block for our \methodname{}, which extends its capabilities by \textbf{(a)} incorporating frame-level text input and \textbf{(b)} enabling the joint generation of both motion and text.
MDM aims to synthesize human motion sequences, denoted as \(\x^{1:N}\), where \(N\) is the length of the sequence. The synthesis process is guided by a sequence-level text condition \(c\), meaning the entire motion sequence is described by a single text prompt. In cases of unconditioned motion generation, the condition is represented as \(c=\emptyset\).

Diffusion is modeled as a Markov noising process, where \( t=0 \) represents the timestep corresponding to the clean data and \( t=T \) corresponds to the fully corrupted data. The samples generated during this process are denoted as \(\{\x^{1:N}_{t}\}_{t=0}^T\), with \( \x^{1:N}_{0} \) being drawn from the data distribution. The transition between steps is defined by:

\begin{equation}
\label{eq:gaussian_noise_process}
q(\x^{1:N}_{t} | \x^{1:N}_{t-1}) = \mathcal{N}(\sqrt{\alpha_{t}}\x^{1:N}_{t-1},(1-\alpha_{t})\mathbf{I}).
\end{equation}

where \(\alpha_{t} \in (0,1)\) indicates the noise level, with $\alpha_i=1 - \beta_i$ and $\beta_i$ being the noise schedule. We drop the sequence length and use \(\x_{t}\) to denote the full sequence at noising step \(t\) for simplicity.
The reverse diffusion process gradually denoises the noisy sequence \(\x_{T}\), with the conditioned motion generation modeling the distribution \( p(\x_{0} | c) \). The denoised data is directly predicted using a model \( G \), where \( \hat{\x}_{0} = G(\x_t, t, c) \) \cite{ramesh2022hierarchical}.

To adapt the diffusion model for human motion, we follow~\cite{Guo2022CVPR} to parameterize the human motion as a 263 dimensional vector.
Due to its redundancy inherent in the motion representation, a simple training objective~\cite{Ho2020DDPM,tevet2023human} can be used, minimizing the expected distance between the original noisy motion \(\x_0\) and the predicted motion \(\hat{\x}_{0}\):
\begin{equation}
\mathcal{L}_\text{simple} = E_{\x_0 \sim q(\x|c), t \sim \mathcal{U}\{1,...,T\}}\| \x_0 - G(\x_t, t, c)\|_2^2.
\label{eq:mdm_loss}
\end{equation}
Notably, this simple loss automatically includes the geometry losses terms described by \cite{tevet2023human}, enforcing physical plausibility and preventing artifacts.

\section{\methodname{}: Unifying Motion Synthesis and Understanding}
\label{sec:method}

In this section, we introduce \methodname, a unified model for joint motion synthesis and understanding, including hierarchical control via text. \methodname{} generates high-quality motion and text outputs, either from full noise or given conditional inputs such as frame-level text, sequence-level text, a motion sequence, or any subsets thereof,  (see Fig. \ref{fig:overview}) spanning a variety of applications treated in isolation by related works.

To achieve this, our model advances prior single-modality motion diffusion models (see Subsec. \ref{subsec:method_mmd}) to encompass multi-modal distributions, specifically motion, and fine-grained text. We combine motion sequence and fine-grained frame-level texts, maintaining the temporal alignment of these two modalities to enable temporal semantic awareness (see Sec. \ref{subsec:method_temporal_tokens}). Unlike previous works, our multi-modality diffusion process supports joint training across datasets with varying annotations (sequence-level and frame-level) (see Subsec. \ref{subsec:method_data_merging}).

\begin{table*}[h]
\centering
\scriptsize %
\resizebox{\textwidth}{!}{ %
\begin{tabular}{l l l| c c c | c c | c c}
\toprule
\multirow{2}{*}{Method} & \multirow{2}{*}{Training Set} & \multirow{2}{*}{Input} & \multicolumn{3}{c|}{Per-crop semantic correctness} & \multicolumn{2}{c|}{Per-crop Realism} & \multicolumn{2}{c}{Per-seq Realism} \\
& & & R-Prec@3 $\uparrow$ & M2T $\uparrow$ & M2M $\uparrow$ & FID $\downarrow$ & Diversity $\rightarrow$ & FID $\downarrow$ & Diversity $\rightarrow$ \\
\midrule
GT & - & - & $0.735^{\pm 0.008}$ & $0.663^{\pm 0.000}$ & $1.000^{\pm 0.000}$ & $0.000^{\pm 0.000}$ & $1.375^{\pm 0.005}$ & $0.000^{\pm 0.000}$ & $1.391^{\pm 0.003}$ \\
\midrule
TEACH & BABEL & f & $0.588^{\pm 0.007}$ & $0.623^{\pm 0.001}$ & $0.575^{\pm 0.000}$ & $0.155^{\pm 0.001}$ & $1.340^{\pm 0.003}$ & $0.304^{\pm 0.001}$ & $1.344^{\pm 0.003}$ \\
DoubleTake & BABEL & f & $0.544^{\pm 0.013}$ & $0.602^{\pm 0.002}$ & $0.560^{\pm 0.001}$ & $0.195^{\pm 0.002}$ & $1.332^{\pm 0.005}$ & $0.353^{\pm 0.002}$ & $1.337^{\pm 0.004}$ \\
STMC & HML & f & $0.528^{\pm 0.012}$ & $0.599^{\pm 0.000}$ & $0.616^{\pm 0.010}$ & $0.156^{\pm 0.000}$ & $1.358^{\pm 0.005}$ & $0.233^{\pm 0.000}$ & $1.362^{\pm 0.005}$ \\
FlowMDM & BABEL & f & $0.618^{\pm 0.007}$ & $0.631^{\pm 0.002}$ & $0.652^{\pm 0.001}$ & $0.101^{\pm 0.001}$ & $1.352^{\pm 0.006}$ & $0.211^{\pm 0.002}$ & $1.375^{\pm 0.005}$ \\
Ours & BABEL & f & $0.636^{\pm 0.017}$ & $0.633^{\pm 0.004}$ & $0.677^{\pm 0.002}$ & $0.087^{\pm 0.002}$ & $1.366^{\pm 0.009}$ & $0.180^{\pm 0.004}$ & $1.374^{\pm 0.002}$ \\
Ours & HML$\cap$
BABEL & f & $\underline{0.668}^{\pm 0.009}$ & $\underline{0.643}^{\pm 0.002}$ & $\underline{0.698}^{\pm 0.002}$ & $\underline{0.071}^{\pm 0.001}$ & $\underline{1.372}^{\pm 0.005}$ & $\underline{0.150}^{\pm 0.001}$ & $\underline{1.378}^{\pm 0.003}$ \\
Ours & HML$\cap$
BABEL & f + s & $\boldsymbol{0.679}^{\pm 0.006}$ & $\boldsymbol{0.644}^{\pm 0.001}$ & $\boldsymbol{0.706}^{\pm 0.002}$ & $\boldsymbol{0.066}^{\pm 0.002}$ & $\boldsymbol{1.373}^{\pm 0.009}$ & $\boldsymbol{0.133}^{\pm 0.004}$ & $\boldsymbol{1.381}^{\pm 0.006}$ \\
\bottomrule
\end{tabular}
}
\vspace{-2mm}
\caption{\textbf{Frame-Level to Text evaluation.} \textit{Per-crop} refers to text segment level evaluation. 
\textit{Training Set} specifies the dataset used for training. \textit{Input} specifies the type of text input. \textit{f}: frame-level text, \textit{s}: sequence-level text. \textit{f+s} demonstrates that combining multi-level conditioning signals can enhance model performance in terms of semantic correspondence. The evaluation is repeated 10 times, and $\pm$ indicates the 95\% confidence intervals.
}
\label{tab:t2m_merged}
\end{table*}

\subsection{Multi-Modal Motion and Text Diffusion}
\label{subsec:method_mmd}
Previous motion synthesis models mainly focus on text-to-motion synthesis tasks~\cite{zhang2022motiondiffuse,tevet2023human,barquero2024seamless,zhou2024avatargpt,jiang2024motiongpt,petrovich2024multi,shafir2023human,zhang2024finemogen,zhang2024motion,TEACH:3DV:2022,xie2024omnicontrol, motionlcm}. Some recent methods also generate sequence-level text descriptions~\cite{zhou2024avatargpt, jiang2024motiongpt}, but lack the temporal awareness and alignment we propose with \methodname{}. Moreover, no model currently supports the joint generation of motion and text. 
This motivates our holistic model of motion synthesis and understanding, \methodname{}, working in a multi-modal, joint probabilistic fashion we introduce next.

Similar in spirit to~\cite{bao2023one} that focuses on joint probabilistic modeling of 2D images and text, our method models the distribution of two temporal modalities under global conditioning. More concretely, a frame-level text sequence, $\y^{1:N}$ is denoted analogously to the motion sequence $\x^{1:N}$, where $N$ denotes the sequence length and \(\{\y^{1:N}_{t}\}_{t=0}^T\) are the noise samples created via Eq. \ref{eq:gaussian_noise_process}.
Similarly, we drop the notation of sequence length in the following for simplicity.
With that, multi-modal diffusion can be achieved by extending $G$ to $G_\theta(\mathbf{x}_{t^x}, \mathbf{y}_{t^y}; t^x, t^y)$ with the additional process and including the separately scheduled diffusion timesteps $t^x$, $t^y$ for motion and text respectively.
By virtue of this formulation, the joint distribution $p(\x, \y)$ can be sampled at inference time, starting the denoising process with  $G_\theta(\mathbf{x}_{T}, \mathbf{y}_{T}; T, T)$, and the conditional $p(\x|\y)$ by $G_\theta(\mathbf{x}_{T}, \mathbf{y}; T, 0)$ and analogously $p(\y|\x)$. %
Specifically, we jointly train the model via  
\begin{equation}
    \min_\theta 
      \mathbb{E}_{(\mathbf{x}_0, \mathbf{y}_0), t^x, t^y}
      \mathbb{E}_{\mathbf{x}_{t^x},
        \mathbf{y}_{t^y}}
        \|G_\theta(\mathbf{x}_{t^x}, \mathbf{y}_{t^y}; t^x, t^y, c) - (\mathbf{x}_0, \mathbf{y}_0)\|_2^2
\label{eqn:unidiffuser}
\end{equation}
where $\theta$ are weights parametrizing $G$ and $\mathcal{U}$ is the discrete uniform distribution and expectation is taken over distributions: $(\mathbf{x}_0, \mathbf{y}_0) \sim p(x,y)$, $t^x \sim \mathcal{U}\{0,...,T\}$, $t^y \sim \mathcal{U}\{0,...,T\}$, $\mathbf{x}_{t^x} \sim q(\mathbf{x}_{t^x} | \mathbf{x}_0),
        \mathbf{y}_{t^y} \sim q(\mathbf{y}_{t^y} | \mathbf{y}_0)$.

\subsection{Temporally aligned Text and Motion Encoding}
\label{subsec:method_temporal_tokens}
We find that appropriate architectural integration of two modalities (text and motion) into the joint formulation is a key performance factor.

A simple integration is to treat motion and text as separate modalities and as input to the Transformer.
An even more structure-agnostic approach is proposed by UniDiffuser~\cite{bao2023one}, where each token of their text encoding is fed separately as input.
We find that both these variants lead to performance issues.

In contrast, in our setting of motion and text sequences, we find temporal alignment to be the key.
A simple, yet effective implementation is the concatenation of motion and text into joint encodings along the temporal dimension.
Instead of learning to correlate word positions with motion positions, alignment is directly given through the input encoding.

However, this alone does not guarantee performance.
We encode text into the space of CLIP~\cite{Radford2021LearningTV} with a pertained model.
Using the full encodings of pose and text as token creates issues.
We hypothesize this is due to an excessive capacity spent on the high-dimensional text tokens.
We solve this by projecting CLIP embeddings down to 50 dimensions via PCA~\cite{WOLD198737} and find this improves performance drastically. 
To get back to text labels from embeddings after diffusion, we match the predicted clip embedding to our database of text labels to obtain the output text using the closest match.

\subsection{Data Merging}
\label{subsec:method_data_merging}
The popular AMASS dataset~\cite{amass} of natural human motion, represented by the SMPL body model~\cite{smpl} has recently been annotated in two efforts, namely BABEL~\cite{BABEL:CVPR:2021} and HumanML3D~\cite{Guo2022CVPR}.
While HumanML3D annotations consist of sequence-level text annotation, that is, a single text annotation for a motion clip, the BABEL annotations consist of frame-level annotations, assigning semantic label to the pose for each frame of the motion sequence.
Instead of restricting to use one at a time, as in prior works, \methodname{} is directly trained on both jointly, using sequence level HumanML3D annotations as condition $c$ and frame level sequences as $\y^{1:N}$.

A challenge however lies in that both datasets annotate different subsets of AMASS.
A trivial solution is to consider overlapping annotations of motions.
We denote our model trained with this scenario \methodname{} 
 \textit{overlap}, and investigate the performance in experiments (see Sec. \ref{sec:experiments}).

\begin{figure*}
    \centering
    \includegraphics[width=\linewidth]{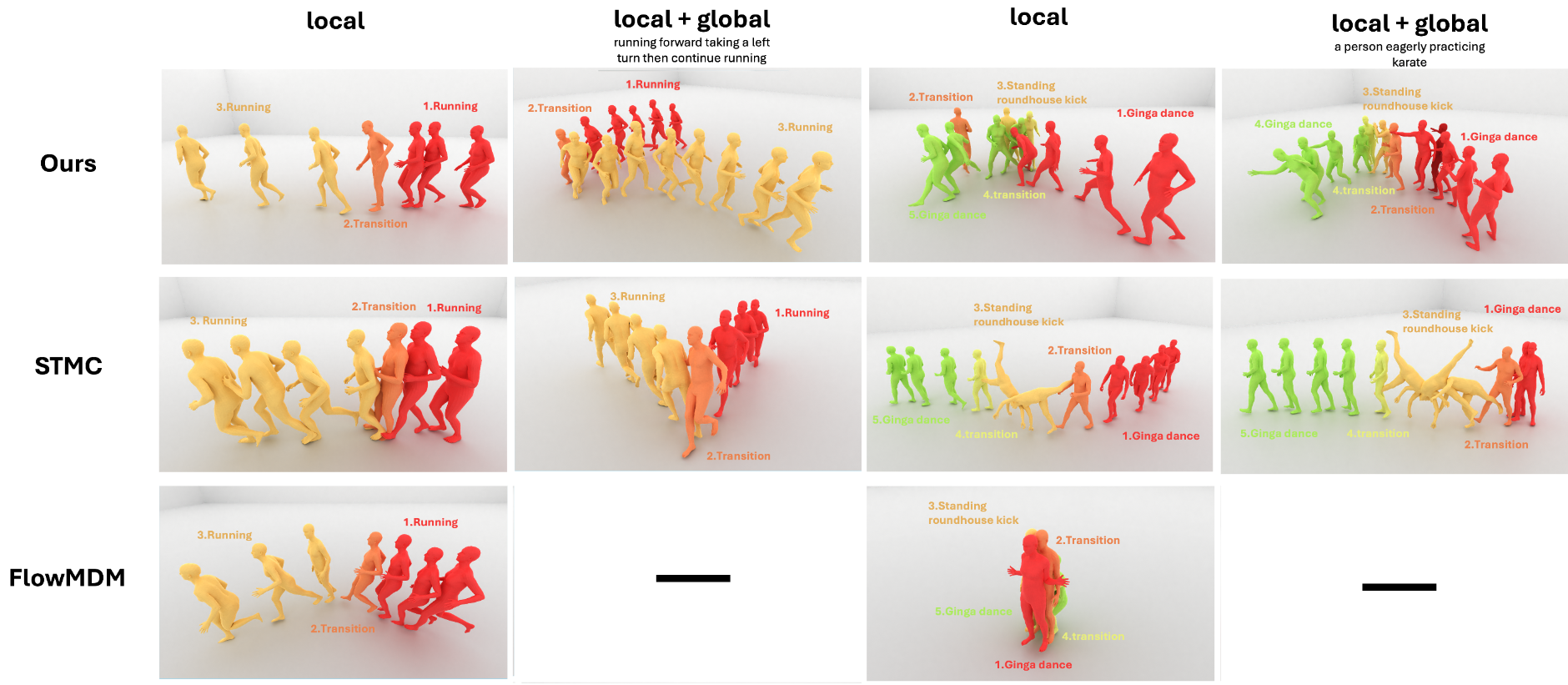}
    \vspace{-5mm}
    \caption{\textbf{Text2Motion qualitative results.} \textbf{Columns 1,3}: Local text is the input to our method and baselines STMC~\cite{petrovich2024multi} (adapted) FlowMDM~\cite{barquero2024seamless}. 
    \textbf{Columns 2, 4}: Both local and global text are the input our method and STMC.
    Our model performs well regardless of the complexity of the local text, in contrast to STMC which fails to generate Ginga dance in columns 3 and 4 and performs walking instead. FlowMDM cannot be conditioned on both global+local text.
    \label{fig:text2motion}
    }
\end{figure*}

\section{Experiments}
\label{sec:experiments}
In this section we investigate the benefit of hierarchical text at inference and training time (i.e. usage of frame-level and sequence-level text).
We show the versatility of \methodname{}'s unification of synthesis and understanding.
Specifically, allowing for frame-level text to motion (Subsec. \ref{subsec:text2motion_frame}) and we for the first time show motion-to-frame-level text (Subsec. \ref{subsec:motion2text}), including a real-world application scenario.
Finally, \methodname{} is the first model to show joint generation of motion along with frame-level understanding (Subsec. \ref{subsec:joint_generation}).
In the ablation study, we show that the proposed multi-modality strongly improves generation quality compared to our backbone MDM~\cite{tevet2023human}. (Subsec. \ref{subsec:ablation})

\paragraph{Implementation Details.} We utilize a temporally aware transformer, similar to MDM~\cite{tevet2023human}. Text inputs are encoded using pretrained CLIP, followed by PCA reduction. Our model is trained on a single A100, with training spanning approximately 40 hours. Please refer to \ref{subsec:method_data_merging} for details on training data.

\paragraph{Baselines}
We compare our model to the publicly released works that are capable of frame-level text-to-motion generation: auto-regressive model \textbf{TEACH}~\cite{TEACH:3DV:2022}, \textbf{DoubleTake}~\cite{shafir2023human} based on diffusion sampling, \textbf{FlowMDM}~\cite{barquero2024seamless}, a diffusion model based on Blended Positional Encoding and \textbf{STMC}~\cite{petrovich2024multi}, a post-hoc test time method stichting individual predictions of MDM~\cite{tevet2023human}. 
Note that neither Teach, FlowMDM nor STMC supports hierarchical training. 
Since STMC admits overlapping control signals 
we compare to it in terms of hierarchical control. 
Since no prior works allow for training on sequence and frame-level text input, models are either trained on BABEL(frame-level) or HumanML3D(sequence-level) data, as indicated in our result tables.
Please refer to our supp. document for more details.
\\
\paragraph{Evaluation Metrics}
First, we introduce our \textbf{semantic metrics}, measuring how well the generated motions correspond to their text descriptions. \textbf{R-Precision}~\cite{Guo2022CVPR} assesses the accuracy of ranking the correct ground-truth text corresponding to a predicted motion at the top positions (Top-1, Top-2, and Top-3) within a set that includes 32 randomly sampled incorrect text matches. With \textbf{M2T}~\cite{petrovich2024multi}, we measure how well the per-crop motion matches their textual description, we calculate their cosine similarity in the joint-embedding space of TMR++~\cite{bensabath2024cross}. Similarly, the \textbf{M2M}~\cite{petrovich2024multi} score is the cosine similarity between the generated and the ground-truth motion embeddings. %

With our \textbf{realism metrics}, we measure how well the generated motion distributions fit the ground truth one.
We utilize the \textbf{Frechet Inception Distance (FID)}~\cite{FID} to measure the distribution distances and \textbf{Diversity}~\cite{Guo2022CVPR} computes the distributions variance, both in the TMR++ as the embedding space.

\begin{table*}[h]
\centering
\scriptsize %
\resizebox{\textwidth}{!}{ %
\begin{tabular}{l l l c c c c c c}
\toprule

Method & Training Set & Input & FID $\downarrow$ & Diversity $\rightarrow$ & R-Prec@1 $\uparrow$ & R-Prec@2 $\uparrow$ & R-Prec@3 $\uparrow$ & M2T $\uparrow$ \\
\midrule

GT & - & - & $0.000^{\pm 0.000}$ & $1.391^{\pm 0.003}$ & $0.699^{\pm 0.014}$ & $0.834^{\pm 0.011}$ & $0.878^{\pm 0.005}$ & $0.748^{\pm 0.000}$ \\
\midrule
MDM & HML & s & $0.449^{\pm 0.025}$ & $1.315^{\pm 0.014}$ & $\underline{0.376}^{\pm 0.008}$ & $0.536^{\pm 0.010}$ & $0.639^{\pm 0.010}$ & $0.631^{\pm 0.003}$ \\
Ours & HML-BABEL & f & $\underline{0.152}^{\pm 0.002}$ & $1.377^{\pm 0.006}$ & $0.344^{\pm 0.010}$ & $0.508^{\pm 0.019}$ & $0.587^{\pm 0.007}$ & $0.648^{\pm 0.003}$ \\
Ours & HML-BABEL & s & $0.195^{\pm 0.003}$ & $\underline{1.381}^{\pm 0.011}$ & $0.375^{\pm 0.021}$ & $\underline{0.539}^{\pm 0.018}$ & $\underline{0.655}^{\pm 0.016}$ & $\underline{0.653}^{\pm 0.004}$ \\
Ours & HML-BABEL & f + s & $\boldsymbol{0.133}^{\pm 0.003}$ & $\boldsymbol{1.382}^{\pm 0.002}$ & $\boldsymbol{0.424}^{\pm 0.005}$ & $\boldsymbol{0.593}^{\pm 0.011}$ & $\boldsymbol{0.677}^{\pm 0.011}$ & $\boldsymbol{0.678}^{\pm 0.002}$ \\
\bottomrule
\end{tabular}
}
\vspace{-2mm}
\caption{\textbf{Ablation Study on Sequence-level Text2Motion generation.} In this table, we compare with our backbone model MDM\cite{tevet2023human} to study whether introducing multi-modality helps the motion generation performance. Symbols $\downarrow$, and $\rightarrow$ indicate that lower, or values closer to the ground truth (GT) are better, respectively. The evaluation is repeated 10 times, and $\pm$ indicates the 95\% confidence interval.}
\label{tab:t2m_global}
\end{table*}

\subsection{Frame-Level Text2Motion Results}
\label{subsec:text2motion_frame}
We evaluate the Text2Motion task (Tab. \ref{tab:t2m_merged}), where we investigate the effects of frame-level and sequence-level training data.
Qualitative analysis is presented in Fig. \ref{fig:text2motion}.
When we train our model as FlowMDM (best performing prior work) on frame-level labels of all Babel annotations (Tab. \ref{tab:t2m_merged}, Ours \textit{BABEL}) we observe our \methodname{} to be consistently better but still roughly on par as expected since both models are using a backbone similar to MDM~\cite{tevet2023human}.
The slight improvement can be attributed to the temporal input alignment (see Sec.\ref{subsec:method_temporal_tokens}) and the multi-timestap diffusion training (see Sec. \ref{subsec:method_mmd}). 
Next, we significantly reduce the training dataset size to the subset sequences annotated with both HML (frame-level text) and BABEL (sequence-level text) (cf. Tab. \ref{tab:t2m_merged} Ours HML-BABEL f). Although one could expect a performance decrease, we find the opposite, a strong consistent performance increase in all metrics - suggesting the strong positive impact of multi-model training. Notably, this is the case although only frame level inputs are given for the evaluation and sequence-level inputs only enrich the models training data.
Finally, we investigate the effect of adding sequence-level text into the model for evaluation (cf. Tab. \ref{tab:t2m_merged} Ours HML-BABEL f + s), again showing a consistent improvement.
In conclusion, the evaluation shows cross-modal generalization, consistently improving the results. 

\begin{figure*}
    \centering
    \includegraphics[width=1\linewidth]{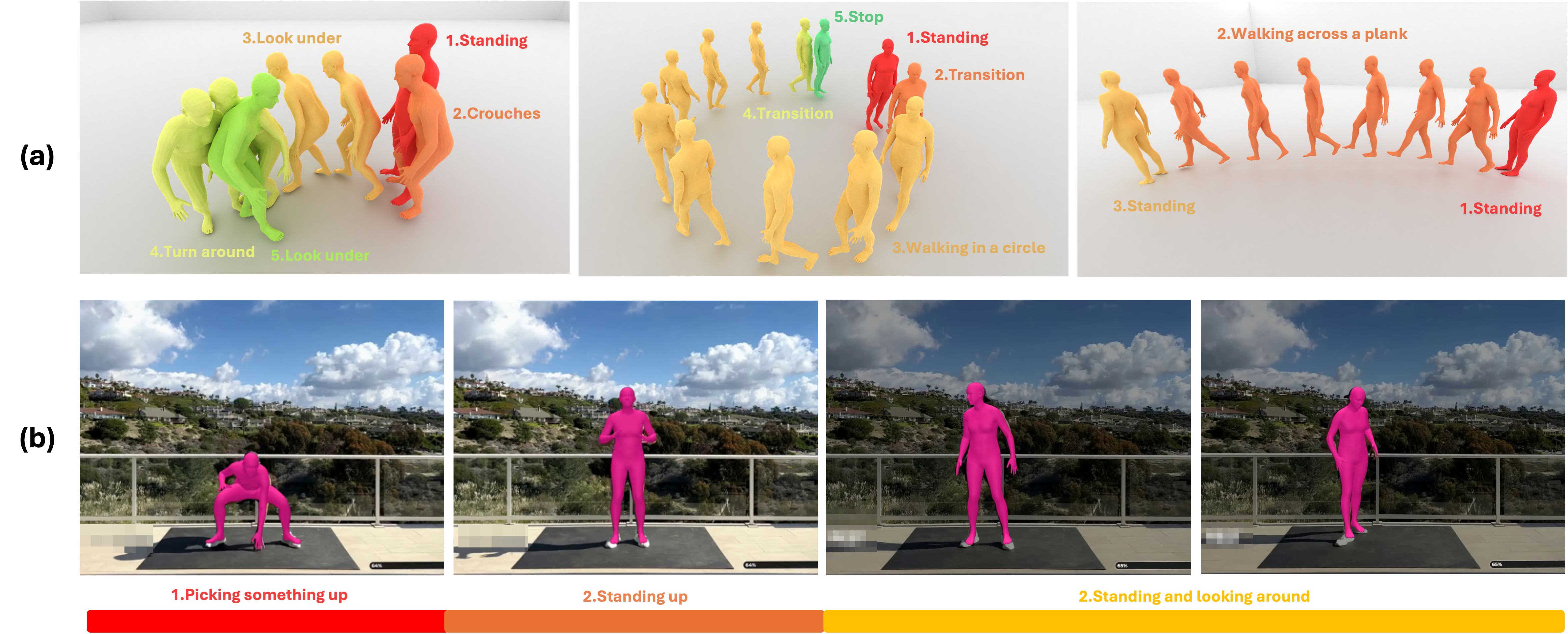}
    \caption{\textbf{Motion2Text understanding of MoCap and YouTube data.} 
    \textbf{(a)} Given an input MoCap sequence, we use \methodname{} to predict frame-level local text.
    \textbf{(b)} We annotate human motion from YouTube videos with frame-level text.
    We lift 2D videos to 3D human motion via frame-by-frame pose estimators~\cite{goel2023humans}. We visualize the SMPL human pose (\textbf{Pink}) overlayed on the YouTube videos frames.
    Then we run \methodname{} to predict frame-level annotations (colored text descriptions below the frames).
    Annotations could serve as valuable audio close captions for the visually impaired.
    }
    \label{fig:motion2text}
\end{figure*}

\subsection{Applications}
Please see these and further results in motion in the supplementary video.

\paragraph{Motion2Text}
\label{subsec:motion2text}
Here we show \methodname{}s capabilities of predicting frame-level text given human motion.
This is a novel task, prior work is not able to do.
We, therefore, restrict ourselves to qualitative evaluations. See Fig. \ref{fig:motion2text}, where we use \methodname{} to annotate MoCap data and Youtube videos with motion descriptions.

\paragraph{Hierarchical Text2Motion:} We show that \methodname{}, although not directly trained for this task, shows generalization capabilities to compositional text conditioning, where global-text and local-text are giving different but complementary conditioning (see. Fig. \ref{fig:teaser}).

\begin{figure*}
    \centering
    \includegraphics[width=\linewidth]{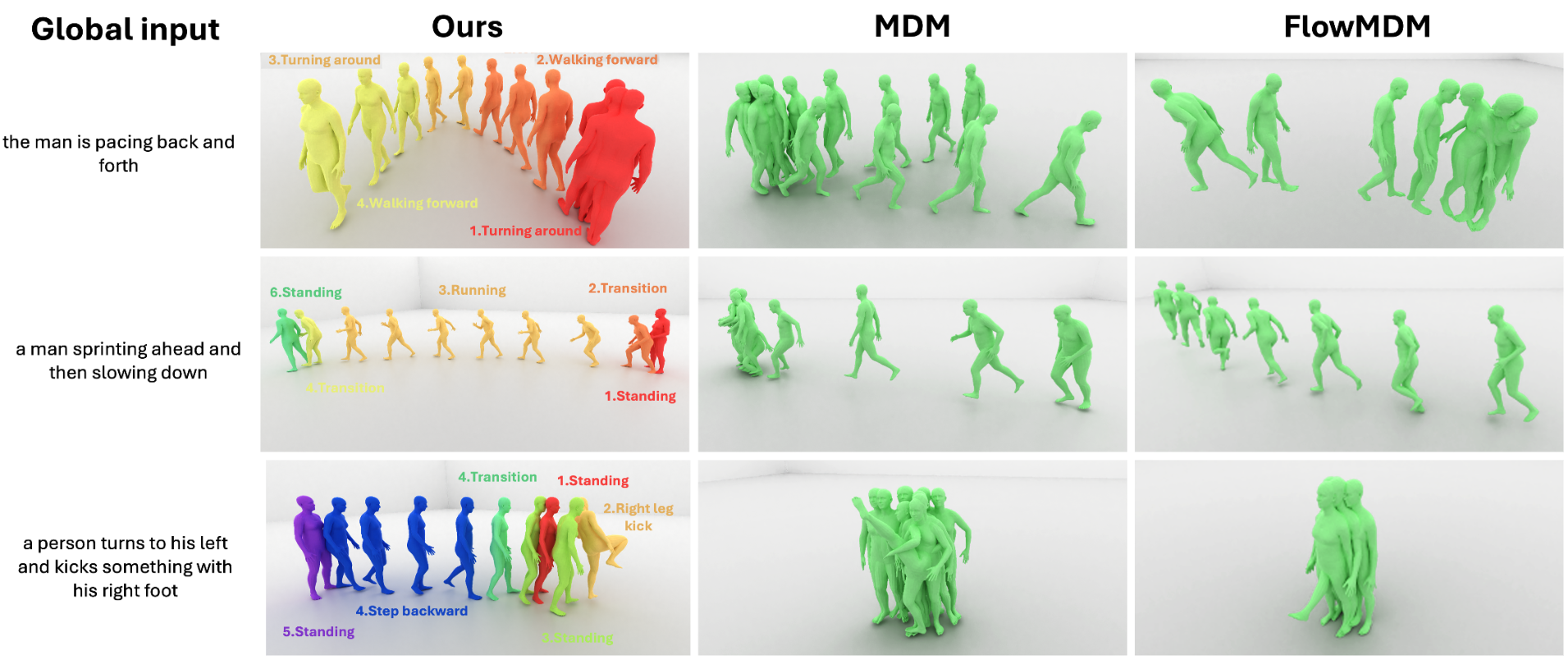}
    \caption{\textbf{Joint text and motion generation results.} 
    Input to the models is only the global text shown on the left. 
    We compare the generated motion of ours, MDM~\cite{tevet2023human} and FlowMDM~\cite{barquero2024seamless}. Our method jointly predicts the frame-level labels, so we can annotate sub-sequences, while MDM and FlowMDM can only generate the motion.}
    \label{fig:joint_generation}
\end{figure*}

\begin{figure*}
    \centering
    \includegraphics[width=\linewidth]{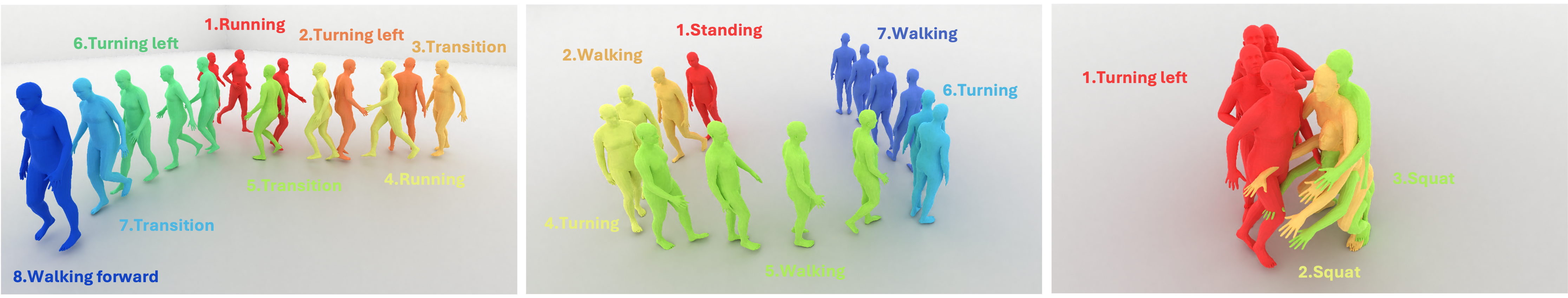}
    \caption{\textbf{Unconditional joint text and motion generation}. Our model, by design, generates poses aligned with local text.}
    \label{fig:joint-generation-no-input}
\end{figure*}

\paragraph{Joint text and motion generation}
\label{subsec:joint_generation}
\methodname{} can jointly generate human motion and corresponding frame-level text, allowing users to not only generate motion but also to directly understand the generated sequence on a frame level. 
Prior work can not perform this task, see Fig. \ref{fig:joint_generation} for conditional joint generation and in Fig. \ref{fig:joint-generation-no-input} for unconditional generation.

\paragraph{Motion Editing for Content Creation}
We show the application of \methodname{} to content creation, where a user specifies a desired motion sequence via rough global text and obtains the motion sequence with a frame-level script. The user succeeds by editing the frame-level script and regenerates the motion to obtain the desired edits, see Fig.~\ref{fig:teaser}.

\subsection{Ablation: Importance of Multi-Modality}
\label{subsec:ablation}
In this section, we investigate the importance of our unification of multiple modalities.

\paragraph{Flexibility.} As seen in previous experiments, this allows to generate high-quality motion and text outputs, either from full noise or given conditional inputs such as frame-level text, sequence-level text, a motion sequence, or any subsets thereof,  (see Fig. \ref{fig:overview}) spanning a variety of applications treated in isolation by related works.

\paragraph{Improved quality.} Additionally, we ablate that the included multi-modality also allows for improved generation quality.
For this, we compare our model trained on multi-modal against our backbone architecture MDM~\cite{tevet2023human}, which does not include frame-level text in output or input, nor is equipped with the flexible multi-modal diffusion.

Our model, used with the same sequence-level text input data (Table \ref{tab:t2m_global}, input: s), as MDM, drastically improves MDM in terms of FID and diversity, but also improves or is on par in other metrics.
Since the backbone transformer is the same, this shows the strength of the proposed multi-modal training.
Notably, this effect is visible even though our training dataset is only a 30\% subset of the MDM training dataset. 

Combining sequence-level and frame-level text (Table \ref{tab:t2m_global}, input: f+s) shows a further significant improvement, improving MDM in all metrics.
This improvement does not stem from the addition of frame-level text input alone since, in isolation, frame-level labels do not achieve this quality (see Table \ref{tab:t2m_global}, input: f). 
We find the interaction between frame-level and sequence-level inputs is the reason for the improvements. 
In conclusion, the proposed multi-modality is the key factor allowing for improved generation quality.

\section{Conclusions}
\label{sec:conclusions}
We introduced \methodname{}, the first unified multi-task human motion model capable of both flexible motion control and frame-level motion understanding. 
Using a flexible multi-model diffusion scheme, \methodname{} solves several tasks in a unified fashion.
Specifically, it unifies tasks that are usually treated in separation by prior works, such as \textit{Frame-Level Text-to-Motion},\textit{ Sequence-Level Text-to-Motion} and \textit{Motion-to-Text}, into a single simple unified model, trained a single time. 
Importantly, \methodname{}'s flexibility also allows for novel tasks not previously considered by prior work like 1.) unconditional generation of human motion with corresponding frame-level text descriptions and 2.) generation of frame-level text from motion, providing granular, time-aware annotations.
We show \methodname{} opens up new applications: 1.) hierarchical control, allowing users to specify motion at different levels of detail,
2.) obtaining motion text descriptions for existing MoCap data or YouTube videos and
3.) allowing for editability, generating motion from text, and editing the motion via text edits. 
Moreover, \methodname{} attained state-of-the-art results for the frame-level text-to-motion task on the established HumanML3D dataset showing the proposed multi-modality is the key factor allowing for improved generation quality.

{\small
\PAR{Acknowledgments:} Special thanks to Xiaohan Zhang for helping with the related work and other RVH and AVG members for the help and discussion. Thanks to Mathis Petrovich, Léore Bensabath, and Prof. Gül Varol for the discussion and helpful information on TMR++\cite{bensabath2024cross}. Prof. Gerard Pons-Moll and Prof. Andreas Geiger are members of the Machine Learning Cluster of Excellence, EXC number 2064/1 - Project number 390727645. Gerard Pons-moll is endowed by the Carl Zeiss Foundation. Andreas Geiger was supported by the ERC Starting Grant LEGO-3D (850533). Julian Chibane is a fellow of the Meta Research PhD Fellowship Program - area: AR/VR Human Understanding.
}

{
    \small
    \bibliographystyle{ieeenat_fullname}
    \bibliography{bibliography}
}

\clearpage
\setcounter{page}{1}
\setcounter{section}{0}
\renewcommand\thesection{\Alph{section}}
\maketitlesupplementary

In the following, we start with the supplementary video in Sec.~\ref{sec:video} and discuss the details of training data in Sec.~\ref{sec:trainingdata}. Then, we present the details of our evaluation setup in Sec.~\ref{sec:evalsetup}, followed by implementation details in Sec.~\ref{sec:more_implementation_details}, additional results in Sec.~\ref{sec:more_results} and Sec. ~\ref{sec:more_apps}. Finally, we demonstrate our model's advantage over LLMs and other motion-to-text models in Sec. ~\ref{sec:m2t_baselines}.

\section{Video with Qualitative Results}\label{sec:video}

We provide videos to further explain our method and to present the results with animated motions, showing a clearer comparison across various tasks and against other baselines. Supplementary results can be found in the accompanying ZIP file.

\section{Training Data}\label{sec:trainingdata}

\begin{figure}
    \centering
    \includegraphics[width=\linewidth]{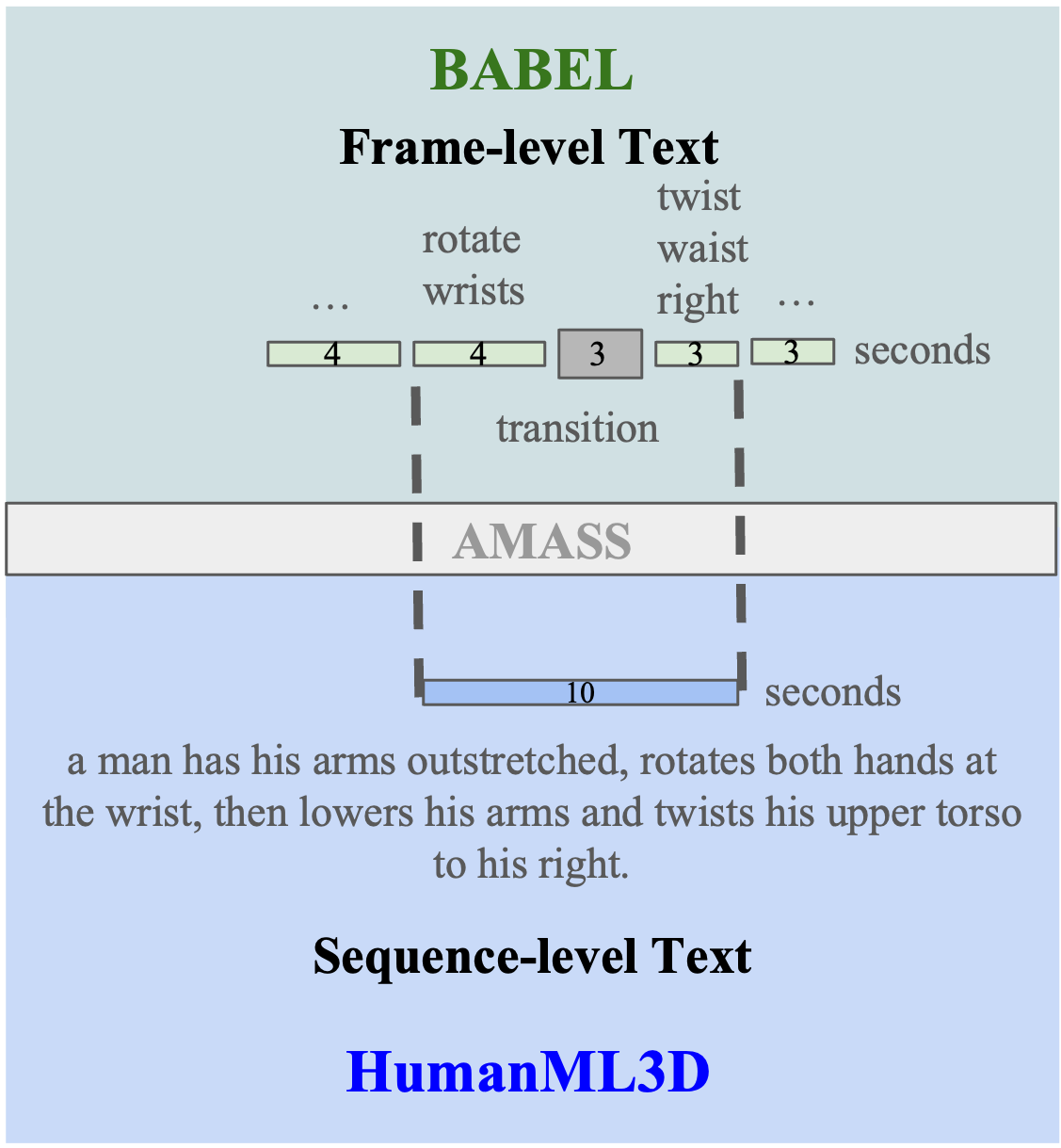}
    \caption{\textbf{We merged HumanML3D and BABEL based on their time correspondence with AMASS.} Each sequence (approximately 1-10 seconds) in HumanML3D includes 3-4 sequence-level annotations in sentence format, as illustrated in the blue area. In contrast, BABEL provides separate annotations for atomic actions with varying lengths, where the text labels are primarily short phrases aligned at the frame level, as shown in the green area.}
    \label{fig:babel_hml}
\end{figure}

\methodname \ is trained on an overlapping subset of BABEL~\cite{punnakkal2021babel} and HumanML3D~\cite{Guo2022CVPR}, utilizing both sequence-level and frame-level text as input. Fig.  \ref{fig:babel_hml} illustrates the data alignment and merging process. However, since these two datasets are independently labeled and cover different subsets of AMASS~\cite{mahmood2019amass}, they do not fully overlap. The overlapping portion comprises only 8,829 motion sequences (excluding left-right flipping), which represents approximately 30.25\% of the HumanML3D dataset (23,384 sequences). This overlapped dataset includes motion sequences, sequence-level text descriptions, and frame-level text descriptions.

\section{Evaluation Setup}\label{sec:evalsetup}

In this section, we outline the details of the evaluation setup and how we run baselines under this setup.

For frame-level text-to-motion generation, we use BABEL frame-level text (in short-phrase format) as conditional input, which is also used as our test-time text input. To ensure a fair comparison with other baselines and to maintain consistency with the training data distribution, we use their pre-trained models on BABEL if available.
However, our model is trained on a subset of the HumanML3D training split, which overlaps with the BABEL test split. 
Consequently, we generate a joint test set, excluding training sequences from both. 
Finally, the test set contains 358 sequences and 998 sub-sequences of motion segments.
Our test, train, and validation split will be made available alongside our code and models upon publication.

\paragraph{TEACH} For TEACH~\cite{TEACH:3DV:2022} we use the pre-trained model supplied by the authors on their website, which was trained on BABEL. Since TEACH can not be applied to text segments with very few frames, we set the minimum size of each evaluation sequence to 8 frames.

\paragraph{PriorMDM} For PriorMDM~\cite{shafir2023human}, we compare DoubleTake with our method. To fairly compare DoubleTake with our method, we use the ``Babel\_TransEmb\_GeoLoss'' pre-trained model, as our local text input is based on the BABEL dataset. 
When feeding motion crops into DoubleTake, we specify the length of each motion crop. In DoubleTake's default setup, the handshake size is set to 20 and the blending window size to 10, resulting in a minimum motion crop length of 70. If a motion crop is shorter than 70, the method automatically pads it to this length. However, many motion crops in our test set are shorter than 70, which would cause significant discrepancies between the input and output motion lengths. To maintain similar input and output sizes, we modify the handshake size to 2 and the blending window size to 1. The results under this setup are shown in Table ~\ref{tab:t2m_merged}.

\begin{figure*}
    \centering
    \includegraphics[width=0.85\linewidth]{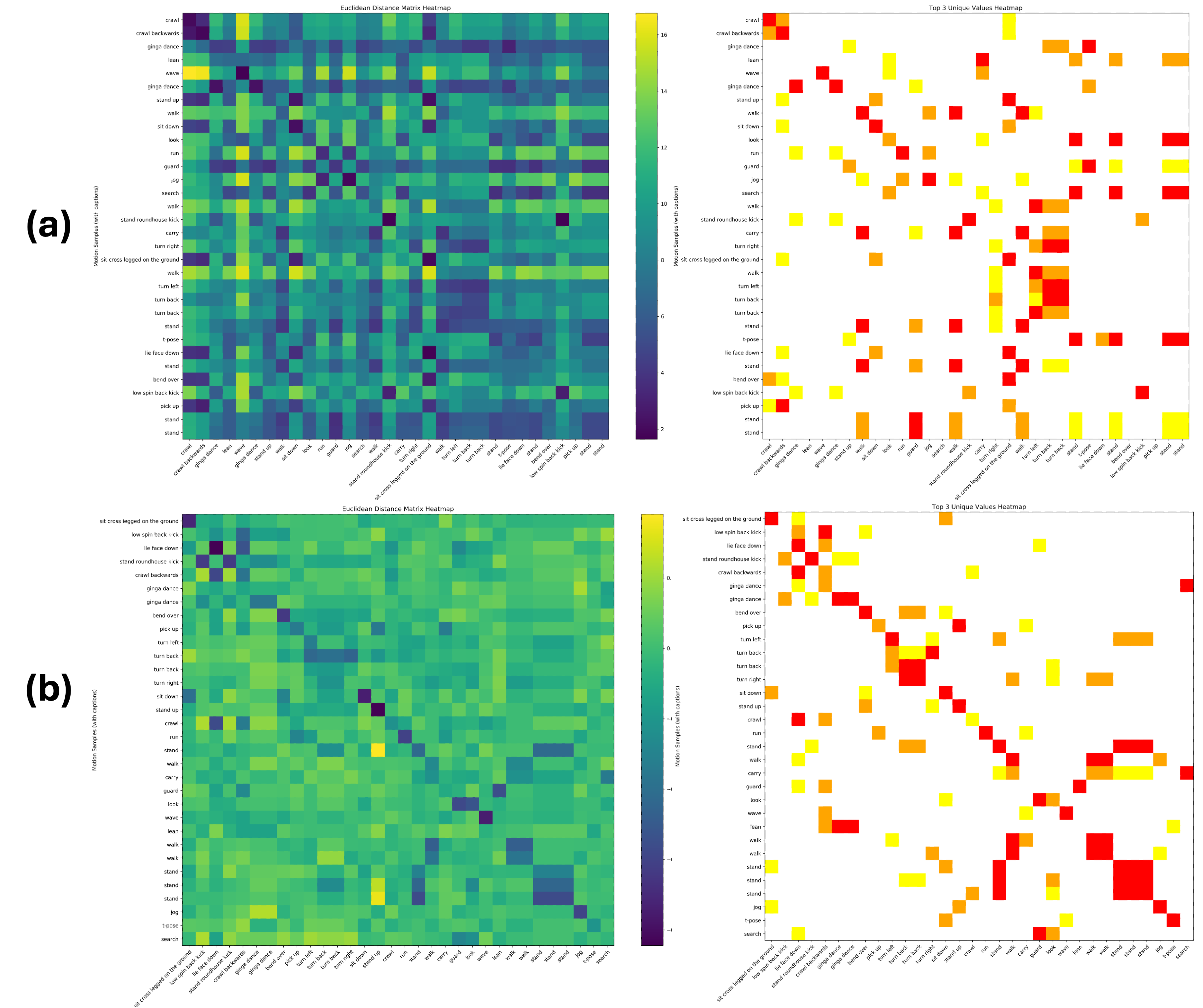}
    \caption{\textbf{The comparison between ground-truth motion-text matching} in the joint embedding spaces of Guo et al.’s model \textbf{(a)} and TMR++ \textbf{(b)}. \textbf{Left:} The heatmap shows the paired motion-text distances, where darker shades indicate smaller distances. The vertical axis represents motion samples, while the horizontal axis represents text samples. \textbf{Right:} The top-3 R-precision scores are displayed for each row, indicating the closest 3 texts to each motion. Red denotes the top 1 match, orange the top 2, and yellow the top 3. If the texts are identical, they are only counted as one.}
    \label{fig:distance_heatmap}
\end{figure*}

\paragraph{STMC} For an entire motion sequence, STMC~\cite{petrovich2024multi} allows specifying the body part for each individual subsequence of motion. To align with our setup, we set the corresponding body part to include all body parts for each motion crop when feeding the motion into STMC.

\paragraph{FlowMDM} To ensure a fair comparison with our method, we use the human motion compositions with the pre-trained BABEL model for the FlowMDM~\cite{barquero2024seamless} method. Since FlowMDM is designed to generate motion compositions seamlessly, there is no need to specify any transition length between atomic motions. Therefore, we directly input the frame-level texts and corresponding lengths, consistent with the input format used for our model.

\paragraph{Evaluation metrics.} For the evaluation metrics—\textbf{Semantic Correspondence} (R-precision, M2T score, M2M score) and \textbf{Realism} (FID, Diversity)—we use TMR++ instead of the commonly used motion and text embedding model from Guo et al.~\cite{Guo2022CVPR}. This choice is driven by the need to evaluate models trained across different datasets and to assess performance at multiple levels of generated motion (per-crop vs. per-sequence). 

For per-crop semantic correctness, we focus on evaluating the alignment of atomic motion crops with their corresponding input text, formatted as BABEL. Additionally, we assess the overall realism of sequence-level motion across crops, which aligns with HumanML3D's sequence-level evaluation. The evaluation model aims to establish a joint latent space for motion and text, performing matching between them based on distance within this shared space.

The commonly used model from Guo et al.~\cite{Guo2022CVPR} is trained solely on HumanML3D. To evaluate BABEL pre-trained models, Shafir et al.~\cite{shafir2023human} retrained this model on BABEL data, and FlowMDM relies on these models for separate evaluations on each dataset. STMC utilizes TMR~\cite{petrovich23tmr}, a retrieval model that demonstrates a better joint latent space compared to the classic evaluation model used by MDM, especially in terms of text-motion distance for ground-truth motion-text pairs. However, TMR is also trained only on HumanML3D, which limits its ability to accurately evaluate both crop-level motions and BABEL text, as well as sequence-level realism.

To address these limitations, we employ the latest model, TMR++~\cite{bensabath2024cross}, which is trained across datasets and delivers highly accurate matching results between ground-truth motion and text, whether in BABEL format (subsequence level, short text phrases) or HumanML3D format (sequence-level, text descriptions in sentences).

For a quantitative comparison, please refer to Table \ref{tab:gt_semantic_metrics}, which evaluates ground-truth motion and text. For qualitative analysis, see Fig. \ref{fig:distance_heatmap}, which presents a heatmap of the matching distance across a random sample of 32 batches.

\begin{table}[h]
\centering
\scriptsize %
\resizebox{\linewidth}{!}{ %
\begin{tabular}{l l c c c c }
\toprule
\multirow{1}{*}{Method} & \multirow{1}{*}{Training Set} & \multicolumn{3}{c}{Per-crop semantic correctness} \\
& &R-Prec@1 $\uparrow$ & R-Prec@2 $\uparrow$ & R-Prec@3 $\uparrow$ \\
\midrule
Guo et al\cite{Guo2022CVPR} & HumanML3D  & $0.281^{\pm 0.005}$ & $0.438^{\pm 0.004}$ & $0.539^{\pm 0.006}$ \\
TMR++\cite{bensabath2024cross} & HumanML3D+BABEL & $\boldsymbol{0.520}^{\pm 0.013}$ & $\boldsymbol{0.659}^{\pm 0.008}$ & $\boldsymbol{0.735}^{\pm 0.008}$ \\
\bottomrule
\end{tabular}
}
\caption{\textbf{Ground-truth matching score comparison across evaluation modals.} In this table, we compare the matching scores across different evaluation models for ground-truth motion and text, averaging over batches of 32 random samples. The results demonstrate that TMR++ is a more reliable model within our evaluation setup.}
\label{tab:gt_semantic_metrics}
\end{table}

\begin{figure*}
    \centering
    \includegraphics[width=0.85\linewidth]{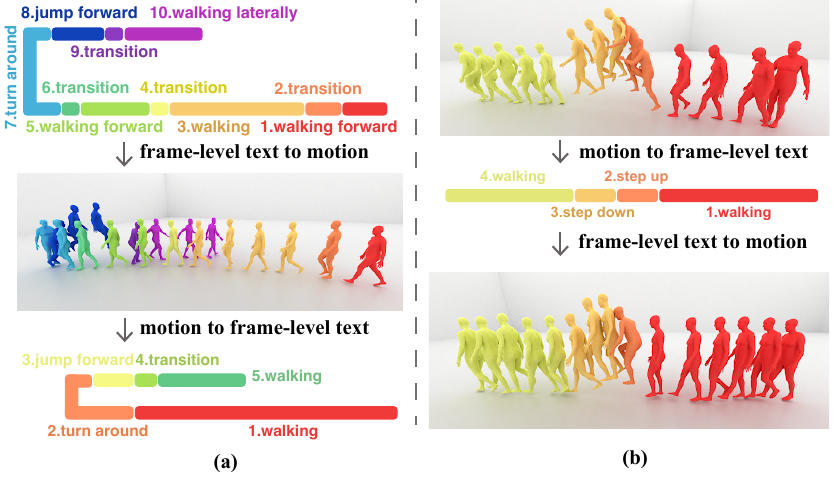}
    \caption{\textbf{Text variation \textbf{(a)} and motion variation \textbf{(b)}} are direct applications that leverage the two conditional distributions modeled by UniMotion.
    Motion variation \textbf{(b)} is achieved by generating frame-level text descriptions from a motion sequence, and then using these descriptions to create a new, semantically similar motion with different content.
    Text variation \textbf{(a)} is produced by reversing this process to create diverse text annotations.}
    \label{fig:chain}
\end{figure*}

\begin{table*}[h]
\centering
\scriptsize %
\resizebox{\textwidth}{!}{ %
\begin{tabular}{l l l c c c c c c}
\toprule
\multirow{2}{*}{Method} & \multirow{2}{*}{Training Set} & \multirow{2}{*}{Input} & \multicolumn{5}{c}{Per-crop semantic correctness} \\
& & & R-Prec@1 $\uparrow$ & R-Prec@2 $\uparrow$ & R-Prec@3 $\uparrow$ & M2T $\uparrow$ & M2M $\uparrow$ \\
\midrule
GT & - & - & $0.520^{\pm 0.013}$ & $0.659^{\pm 0.008}$ & $0.735^{\pm 0.008}$ & $0.663^{\pm 0.000}$ & $1.000^{\pm 0.000}$ \\
\midrule
TEACH & BABEL & f & $0.375^{\pm 0.008}$ & $0.516^{\pm 0.007}$ & $0.588^{\pm 0.007}$ & $0.623^{\pm 0.001}$ & $0.575^{\pm 0.000}$ \\
DoubleTake & BABEL & f & $0.332^{\pm 0.013}$ & $0.467^{\pm 0.013}$ & $0.544^{\pm 0.013}$ & $0.602^{\pm 0.002}$ & $0.560^{\pm 0.001}$ \\
STMC & HML & f & $0.321^{\pm 0.009}$ & $0.452^{\pm 0.012}$ & $0.528^{\pm 0.012}$ & $0.599^{\pm 0.000}$ & $0.616^{\pm 0.010}$ \\
FlowMDM & BABEL & f & $0.389^{\pm 0.009}$ & $0.532^{\pm 0.014}$ & $0.618^{\pm 0.007}$ & $0.631^{\pm 0.002}$ & $0.652^{\pm 0.001}$ \\
Ours & BABEL & f & $0.394^{\pm 0.010}$ & $0.552^{\pm 0.018}$ & $0.636^{\pm 0.017}$ & $0.633^{\pm 0.004}$ & $0.677^{\pm 0.002}$ \\
Ours & HML-BABEL & f & $\underline{0.427}^{\pm 0.011}$ & $\underline{0.587}^{\pm 0.012}$ & $\underline{0.668}^{\pm 0.009}$ & $\underline{0.643}^{\pm 0.002}$ & $\underline{0.698}^{\pm 0.002}$ \\
Ours & HML-BABEL & f + s & $\boldsymbol{0.450}^{\pm 0.018}$ & $\boldsymbol{0.593}^{\pm 0.008}$ & $\boldsymbol{0.679}^{\pm 0.006}$ & $\boldsymbol{0.644}^{\pm 0.001}$ & $\boldsymbol{0.706}^{\pm 0.002}$ \\
\bottomrule
\end{tabular}
}
\vspace{-2mm}
\caption{\textbf{Per-crop semantic correctness evaluation for frame-level Text2Motion generation.}
\textbf{Training Set} specifies the dataset used for training, including BABEL, HumanML3D(HML), or the union/intersection of HML and BABEL. \textbf{Input} specifies the type of text input. \textbf{f}: frame-level text, \textbf{s}: sequence-level text. \textbf{f+s} demonstrates that combining multi-level conditioning signals can enhance model performance in terms of semantic correspondence. Symbols like $\uparrow$ indicates that higher, lower, or values closer to the ground truth (GT) are better, respectively. The evaluation is repeated 10 times, and $\pm$ indicates the 95\% confidence intervals.}
\label{tab:t2m_frame_semantic_all_metrics}
\end{table*}
\vspace{-2mm}

\begin{table*}[h]
\centering
\scriptsize %
\resizebox{\textwidth}{!}{ %
\begin{tabular}{l l l| c c c c | c c c c}
\toprule
\multirow{2}{*}{Method} & \multirow{2}{*}{Training Set} & \multirow{2}{*}{Input} & \multicolumn{4}{c|}{Per-crop Realism} & \multicolumn{4}{c}{Per-seq Realism} \\
 &  &  & FID $\downarrow$ & Diversity $\rightarrow$ & FID\_tmr++ $\downarrow$ & Diversity\_tmr++ $\rightarrow$  & FID $\downarrow$ & Diversity $\rightarrow$ & FID\_tmr++ $\downarrow$ & Diversity\_tmr++ $\rightarrow$ 
\\
\midrule
GT &- &- & $0.000^{\pm 0.000}$ & $8.823^{\pm 0.067}$ & $0.000^{\pm 0.000}$ & $1.375^{\pm 0.005}$ & $0.000^{\pm 0.000}$ & $9.296^{\pm 0.086}$ & $0.000^{\pm 0.000}$ & $1.391^{\pm 0.003}$ 
\\
\midrule
TEACH  & BABEL & f & $2.557^{\pm 0.016}$ & $7.879^{\pm 0.119}$ & $0.155^{\pm 0.001}$ & $1.340^{\pm 0.003}$ & $3.577^{\pm 0.025}$ & $7.605^{\pm 0.066}$ &  $0.304^{\pm 0.001}$ & $1.344^{\pm 0.003}$ 
\\

DoubleTake  & BABEL & f & $2.820^{\pm 0.127}$ & $8.248^{\pm 0.102}$& $0.195^{\pm 0.002}$ & $1.332^{\pm 0.005}$ & $5.619^{\pm 0.268}$ & $7.350^{\pm 0.074}$ & $0.353^{\pm 0.002}$ & $1.337^{\pm 0.004}$ 
\\

STMC  & HML & f & $2.161^{\pm 0.008}$ & $9.250^{\pm 0.130}$ & $0.156^{\pm 0.000}$ & $1.358^{\pm 0.005}$ &$1.295^{\pm 0.017}$ & $8.955^{\pm 0.102}$ & $0.233^{\pm 0.000}$ & $1.362^{\pm 0.005}$
\\

FlowMDM & BABEL & f & $0.885^{\pm 0.043}$ & $8.476^{\pm 0.086}$ & $0.101^{\pm 0.001}$ & $1.352^{\pm 0.006}$& $1.028^{\pm 0.060}$ & $8.691^{\pm 0.127}$ & $0.211^{\pm 0.002}$ & $1.375^{\pm 0.005}$ 
\\

Ours & BABEL & f & $1.206^{\pm 0.079}$ & $\underline{9.007}^{\pm 0.141}$ & $0.087^{\pm 0.002}$ & $1.366^{\pm 0.009}$ &$0.791^{\pm 0.091}$ & $8.899^{\pm 0.159}$ & $0.180^{\pm 0.004}$ & $1.374^{\pm 0.002}$
\\

Ours & HML-BABEL & f & $\underline{0.506}^{\pm 0.024}$ & $\boldsymbol{8.979}^{\pm 0.095}$ & $\underline{0.071}^{\pm 0.001}$ & $\underline{1.372}^{\pm 0.005}$  &  $\underline{0.401}^{\pm 0.030}$& $\underline{8.956}^{\pm 0.123}$ &$\underline{0.150}^{\pm 0.001}$ & $\underline{1.378}^{\pm 0.003}$  
\\

Ours & HML-BABEL & f + s  & $\boldsymbol{0.487}^{\pm 0.021}$ & $9.040^{\pm 0.118}$ & $\boldsymbol{0.066}^{\pm 0.002}$ & $\boldsymbol{1.373}^{\pm 0.009}$ & $\boldsymbol{0.299}^{\pm 0.023}$ & $\boldsymbol{8.978}^{\pm 0.095}$ & $\boldsymbol{0.133}^{\pm 0.004}$ & $\boldsymbol{1.381}^{\pm 0.006}$ 
\\
 
\bottomrule
\end{tabular}
}
\caption{\textbf{Frame-level Text2Motion generation per-crop and per-sequence realism evaluation.} Crop-level realism measures the metrics within each atomic crop, while Seq-level realism measures the fidelity of the overall motion. Symbols $\downarrow$, and $\rightarrow$ indicate that lower, or values closer to the ground truth (GT) are better, respectively.}
\label{tab:t2m_frame_realism_all_metrics}
\end{table*}

\section{Implementation Details}\label{sec:more_implementation_details}

We provide more details about the implementation of our model. We extend the MDM~\cite{tevet2023human} framework to separate time steps for motion and frame-level text, and adjust the input to accept the temporal alignment of both the motion vector and text embedding vector. The model is retrained from scratch using the merged overlapping dataset, with hyperparameters consistent with those suggested by Tevet et al.~\cite{tevet2023human}.

For frame-level text, we use the same CLIP model as used in MDM to generate embeddings. We then applied PCA to condense the dimensionality from 256 to 51, preserving approximately 70\% of the original variance. Our model predicts both the clean motion and the condensed CLIP embeddings for the frame-level texts. To output the texts, we use K-nearest neighbors (KNN) to match the output CLIP embeddings in a pre-computed database. This approach effectively matches nearby CLIP embeddings to the corresponding closest text even with a small variance.

For the training and sampling algorithm, please refer to Algorithm \ref{alg:train}, \ref{alg:sample_one}, \ref{alg:sample_joint}.

\begin{algorithm}[H]
\caption{Training}
\begin{algorithmic}[1]
\Repeat
    \State $\x_0, \y_0, c \sim q(\x_0, \y_0, c)$
    \State $c = \varnothing$ \text{ with probability 10\%} 
    \State $t^x, t^y \sim \text{Uniform}(\{1, 2, \ldots, T\})$
    \State $\epsilon^x, \epsilon^y \sim \mathcal{N}(\mathbf{0}, \mathbf{I})$
    \State Let $\x_{t^x} = \sqrt{\overline{\alpha}_{t^x}} \x_0 + \sqrt{1 - \overline{\alpha}_{t^x}} \epsilon^x$
    \State Let $\y_{t^y} = \sqrt{\overline{\alpha}_{t^y}} \y_0 + \sqrt{1 - \overline{\alpha}_{t^y}} \epsilon^y$
    \State Take gradient step on $\nabla_\theta \|\epsilon_\theta(\x_{t^x}, \y_{t^y}, t^x, t^y, c) - [\x_0, \y_0]\|_2^2$
\Until{converged}
\end{algorithmic}
\label{alg:train}
\end{algorithm}

\begin{algorithm}[H]
\caption{Sampling $\x_0$ conditioned on $\y_0$ (similar for sampling $\y_0$ conditioned on $\x_0$, with or without conditioning on $c$.}
\begin{algorithmic}[1]
\State $\x_0^T \sim \mathcal{N}(\mathbf{0}, \mathbf{I})$
\State $c = \varnothing$ or user specify
\For{$t = T, \ldots, 1$}
    \State $\epsilon \sim \mathcal{N}(\mathbf{0}, \mathbf{I})$ 
    \State $\x_0^{t-1} = \epsilon_\theta^x(\sqrt{\overline{\alpha}_{t^x}} \x_0^{t} + \sqrt{1 - \overline{\alpha}_{t^x}} \epsilon, \y_0, t, 0, c)$
\EndFor
\State \Return $\x_0$
\end{algorithmic}
\label{alg:sample_one}
\end{algorithm}

\begin{algorithm}[H]
\caption{Joint sampling of $\x_0, \y_0$ (with or without condition on $c$)}
\begin{algorithmic}[1]
\State $\x_0^T, \y_0^T \sim \mathcal{N}(\mathbf{0}, \mathbf{I})$
\State $c = \varnothing$ or user specify
\For{$t = T, \ldots, 1$}
    \State $\epsilon^x, \epsilon^y \sim \mathcal{N}(\mathbf{0}, \mathbf{I})$ 
    \State $\x_0^{t-1}, \y_0^{t-1} = \epsilon_\theta(\sqrt{\overline{\alpha}_{t^x}} \x_0^{t} + \sqrt{1 - \overline{\alpha}_{t^x}} \epsilon^x, \sqrt{\overline{\alpha}_{t^y}} \y_0^{t} + \sqrt{1 - \overline{\alpha}_{t^y}} \epsilon^y, t, t, c)$
\EndFor
\State \Return $\x_0, \y_0$
\end{algorithmic}
\label{alg:sample_joint}
\end{algorithm}

\section{More Experiment Results}\label{sec:more_results}

We provide only a subset of the metrics for semantic correspondence and motion realism in the main paper due to space constraints. Here, we provide the complete evaluation.

\begin{figure*}
    \centering
    \includegraphics[width=\linewidth]{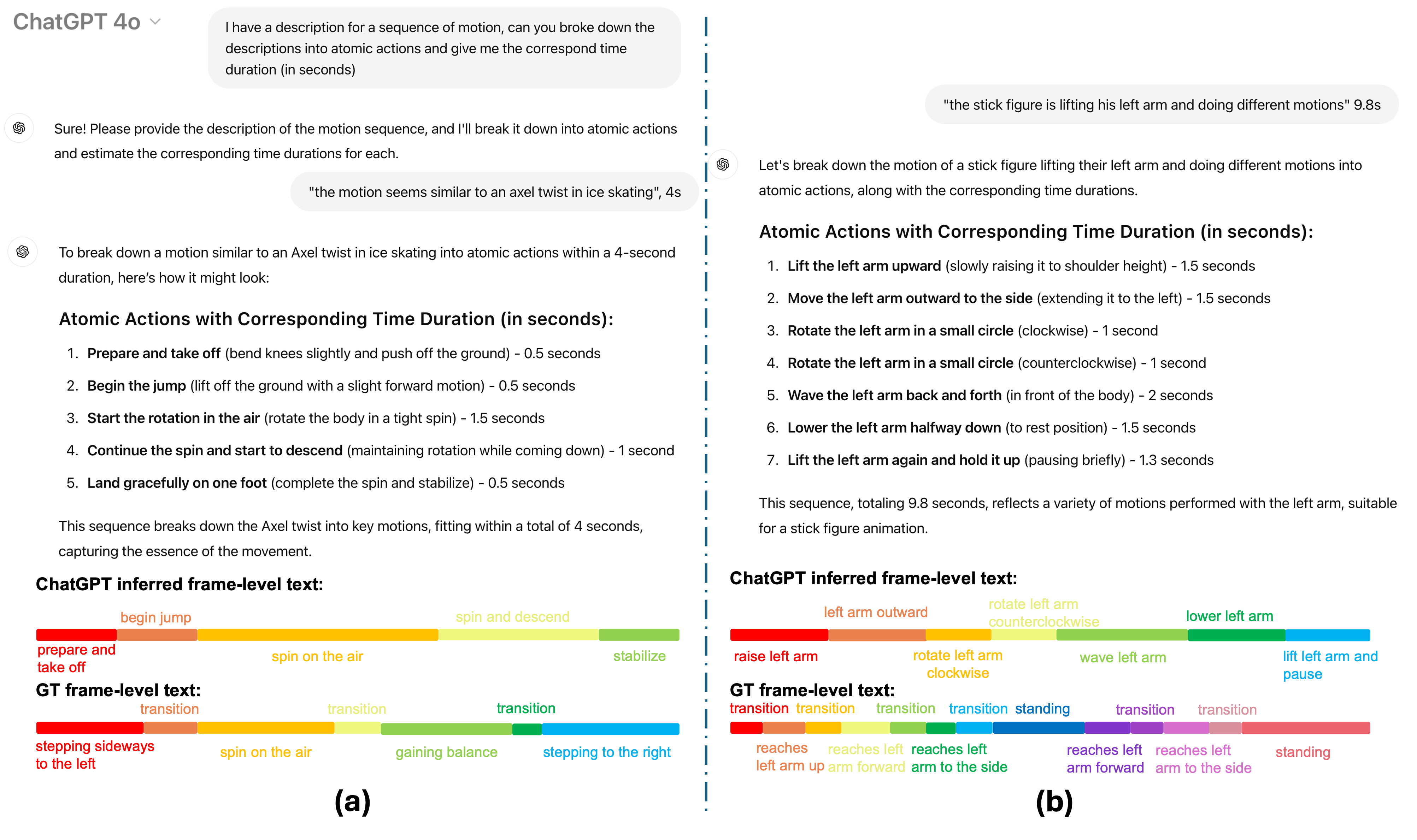}
    \caption{\textbf{Fine-grained motion understanding with LLM.} ChatGPT-4o is used to break down the ground-truth global descriptions into atomic motion and durations. However, there is no alignment between text and motion since the model doesn't take the motion as input. }
    \label{fig:chatgpt}
\end{figure*}

\begin{figure*}
    \centering
    \includegraphics[width=\linewidth]{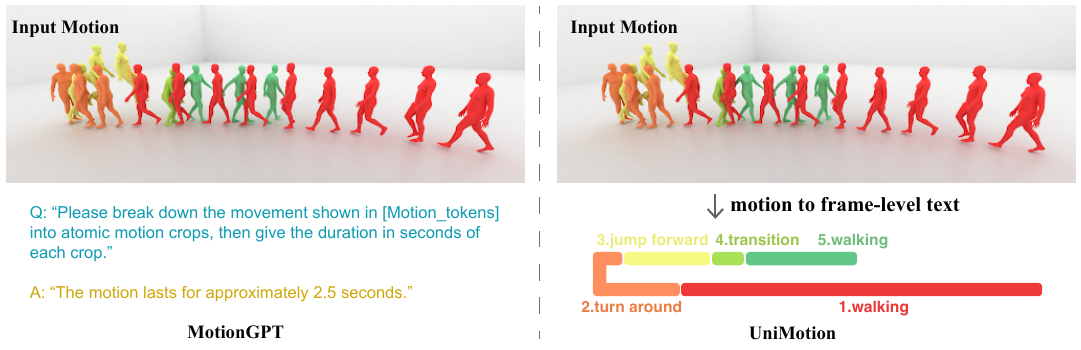}
    \caption{\textbf{Motion understanding comparison with MotionGPT~\cite{jiang2024motiongpt}.} MotionGPT is capable of performing multiple tasks, including motion captioning and question answering. We tasked both MotionGPT (left) and Unimotion (right) with understanding an input motion by breaking it down into motion segments. However, due to MotionGPT's lack of temporal awareness, it was unable to successfully complete this task. Specifically, instead of answering with multiple motion segments, it just predicts an incorrect length for the whole sequence (A: “The motion lasts for approximately 2.5 seconds.”). In contrast, our model is the first to understand motion both semantically and temporally.}
    \label{fig:mgpt}
\end{figure*}

\paragraph{Semantic correspondence}. Tab. \ref{tab:t2m_frame_semantic_all_metrics} lists all three R-precision scores, demonstrating that our method outperforms all baseline methods. These results are consistent with our conclusions in the experiment section of the main paper.

\paragraph{Realism} Tab. \ref{tab:t2m_frame_realism_all_metrics} includes FID and Diversity scores calculated using the evaluation model from Guo et al.~\cite{Guo2022CVPR} for reference. Note that at the crop level, this model provides less stable evaluations because it was trained only on HumanML3D, which contains only squence-level motions. Consequently, FID and Diversity scores from TMR++ offer a more reliable assessment. At the sequence level, both evaluation models yield consistent results. For simplicity and consistency, the main paper presents only FID\_TMR++ and Diversity\_TMR++.

\section{More Applications}\label{sec:more_apps}
Due to space limitations, we only present part of applications in the main paper. Here, we showcase two additional applications that are made possible exclusively by our multi-task model.
Similar to UniDiffuser~\cite{bao2023one}, \methodname  naturally supports applications such as motion variation and text variation. For \textbf{motion variation}, given a motion sequence, we first perform the motion-understanding task to generate frame-level text descriptions aligned with the motion. We then use this frame-level text as input for text-to-motion generation, resulting in a new motion that retains similar semantics but with different content. For \textbf{text variation}, we reverse the process to produce fine-grained text annotation variance. Figure \ref{fig:chain} provides examples of both motion and text variation. For animated results, please refer to the attached videos.

\section{Motion-to-text Understanding Baselines}\label{sec:m2t_baselines}

To establish baselines for our frame-level motion understanding sub-task, we initially attempted to use a large language model (LLM), ChatGPT, to decompose sequence-level inputs and assess potential outputs. However, due to the LLM's lack of motion awareness, the outputs were unreliable when the sequence-level information was vague or incomplete. Even with detailed sequence-level descriptions, the LLM struggled to generate accurate timestamps due to the absence of motion data. Please refer to Fig. \ref{fig:chatgpt} for more details.

We then considered using LLM-based motion models like MotionGPT~\cite{jiang2024motiongpt}, which can process both motion data and text prompts (to request timestamps and atomic text labels). Despite this, MotionGPT also failed in this task. See Fig. \ref{fig:mgpt} for further information.

\end{document}